\documentclass[a4paper,twoside]{article}
\usepackage[utf8]{inputenc}
\usepackage{epsfig}
\usepackage{cite}
\usepackage{subcaption}
\usepackage{calc}
\usepackage{amssymb}
\usepackage{amstext}
\usepackage{amsmath}
\usepackage{amsthm}
\usepackage{multicol}
\usepackage{xcolor}
\usepackage{pslatex}
\usepackage{apalike}

\usepackage{hyperref}
\usepackage{SCITEPRESS} 

\begin{document}
\title{Towards Lightweight Neural Animation : Exploration of Neural Network Pruning in Mixture of Experts-based Animation Models}

\author{\authorname{Antoine Maiorca\sup{1}, Nathan Hubens\sup{1,2}, Sohaib Laraba\sup{1} and Thierry Dutoit\sup{1}}
\affiliation{\sup{1}ISIA Lab, Faculty of Engineering, University of Mons, Mons, Belgium}
\affiliation{\sup{2}Artemis, TelecomSud Paris, Paris, France}
\email{\{antoine.maiorca, nathan.hubens, sohaib.laraba, thierry.dutoit\}@umons.ac.be}
}

\keywords{Neural Character Animation, Neural network pruning}

\abstract{In the past few years, neural character animation has emerged and offered an automatic method for animating virtual characters. Their motion is synthesized by a neural network. Controlling this movement in real time with a user-defined control signal is also an important task in video games for example. Solutions based on fully-connected layers (\textit{MLPs}) and \textit{Mixture-of-Experts (MoE)} have given impressive results in generating and controlling various movements with close-range interactions between the environment and the virtual character. However, a major shortcoming of fully-connected layers is their computational and memory cost which may lead to sub-optimized solution. In this work, we apply pruning algorithms to compress an \textit{MLP}-\textit{MoE} neural network in the context of interactive character animation, which reduces its number of parameters and accelerates its computation time with a trade-off between this acceleration and the synthesized motion quality. This work demonstrates that, with the same number of experts and parameters, the pruned model produces less motion artifacts than the dense model and the learned high-level motion features are similar for both.}

\onecolumn \maketitle \normalsize \setcounter{footnote}{0} \vfill

\section{\uppercase{Introduction}}
\label{sec:intro}
Virtual characters animation is a trending topic for video game and movie industry. It is important that the motion generation algorithm can guarantee a plausible and natural synthesis in order to increase  the immersive factor of the movie or the video game. Thus, the computer vision research community has a special interest in the task of motion synthesis. Moreover, the real-time control of the character trajectory and actions may be an important constraint depending on the use case. 

\cite{pfnn} showed that, with the use of a relatively simple neural network architecture based on \textit{Multi-Layer Perceptrons (MLP)}, it is possible to draw several modes of locomotion for a biped character. The motion data needs first to be aligned on a phase vector that represents the timing of the motion cycle. Then, a regression network whose parameters change dynamically according to the phase vector generates the pose-by-pose motion in an autoregressive manner. Finally, a regression network whose weights vary dynamically according to the phase vector generates the pose-by-pose motion in an autoregressive way. \cite{mann} keeps the same regression network as \cite{pfnn} but proposes upstream a \textit{Mixture-of-Experts (MoE)} to generalize this architecture for the motion of quadruped characters whose the different  locomotion gaits are difficult to align on one phase signal.

However, fully-connected layers are computationally expensive in terms of number of parameters and floating point operations (\textit{FLOPs}) compared to other architectures like \textit{Convolutional Neural Network} for example. Once the network is trained, some weights in the neural network may not learn information that significantly contributes to the quality of the generated signal. Thus, the network capacity may be not fully exploited. Moreover, some of recent architectures can be highly greedy in term of hardware resources and may not be suitable for embedded deployment but reducing its number of parameters would be too costly in terms of performance. Neural network pruning is a research area that aims to compress a neural network while keeping its performance intact. It is widely used when there are non-negligible constraints on hardware resources, such as on embedded devices. The idea behind this concept is to remove unnecessary weights and to retrain the pruned neural network. Practically, these are not removed but nullified. This induces that the pruned weight matrices become sparse. The proportion of zero parameters compared to the original network size is termed as sparsity. In this paper, we explore the impact of pruning a \textit{MLP}-and-\textit{MoE} based neural network in the context of neural character animation. The benefits of this method is to reduce the number of the network parameters and thus theoretically accelerate the computation time while minimizing the performance loss. To the best of our knowledge, this work is the first that applied pruning methods in the context of neural animation. 

First, the state of the art on modern data-driven animation techniques as well as on neural network pruning is presented in section \ref{sec:sota}. Then, the experiments performed are described in section \ref{sec:exp} and the related results in section \ref{sec:res}.

\section{\uppercase{Related Work}}
\label{sec:sota}

\subsection{Data-driven Animation}
\label{subsec:dda}
The task of interactive motion generation is a broad field of research focusing on synthesizing a natural motion from control signals. The problematic of real-time pose prediction from a user-defined trajectory belongs to this paradigm. This has been addressed by data-driven methods using machine learning techniques \cite{ml1,joelle1,motionmatching}. More recent works lie in the field of autoregressive models where, along with the controlled trajectory, the next character pose is predicted with the previous ones. More recent models consider deep learning algorithms with time series analysis architectures such as \textit{RNNs} \cite{lstm1,quaternet}. This family of models combined with probabilistic generative methods draw motions that exhibit less artefact than deterministic models \cite{lstm2,Moglow,TrajVAE}. 
\paragraph{}
\cite{pfnn} has shown that real-time controlled locomotion of a bipedal character can be realized thanks to a phase vector representing the timing of the motion cycle and computed from foot contact with the ground, like said in section \ref{sec:intro}, since \textit{MLPs} do not initially extract any correlation in sequential data. This variable is used to cyclically generates the weights of a \textit{MLP}-based regression network that synthesizes, from motion features such as joint positions, velocities and rotations at the frame $t$, the pose as the next frame. However, this system fails when the motion is difficult to align to a single phase vector, such as quadruped locomotion. To tackle this issue, \cite{mann} proposes to use the technique of \textit{MoE} \cite{moe}. This technique is based on $n$~experts models. Each of them is specialized in the realization of a particular subtask of the given problem. The output $y_i$ of each expert is further weighted by $n$ coefficients $\omega_i$ computed by a so-called gating network. It learns to quantify the importance of each expert to resolve the given task. Thus, the final result $y$ takes into account the effect of the $n$ expert models. 
\begin{equation}
    y = \sum_i^n \omega_i y_i
\end{equation}
In the work of \cite{mann}, the gating network computes
coefficients from the skeleton leg features and simulates the effect of the phase function in \cite{pfnn} for quadruped character. In this case, the experts are sets of parameters learned during training that are then blent. Then, this recombination is used as weights of the regression network. This technique has shown encouraging results in the field of interactive motion generation and constitutes the backbone of recent interactive locomotion generation algorithms. \cite{nsm} employed this method so that a character is able to interact adaptively with objects of variable shape defined in the virtual environment. For example, he can sit on a chair or carry a box without having any prior knowledge of their shape. More complex movements than locomotion can also be synthesized. \cite{local_phase} propose a framework to generate basketball movements with awareness of the ball and a direct opponent. The model must therefore learn to analyze contacts other than feet on the ground, such as the hand-ball contact. For this, they introduce the concept of local phase where each phase is linked to a particular subset of the body. The movement is thus no longer aligned on a single global phase but on several phases and they showed that this helps the model to learn multi-contact motion. More recently, \cite{boxe} have succeeded, still with \textit{MoE}, in generating with simple user-defined control signals a variety of martial arts movements while allowing a close-character interaction. \cite{mvae} implemented the method of \cite{mann} with a \textit{VAE} architecture. This solution allows to define the character controller through Reinforcment Learning. Thus, the avatar can be controlled from joystick input, learn to follow a predefined trajectory or even freely explore an environment like a maze.

\subsection{Pruning}

Pruning methods primarily differ according to three factors: (1) the granularity at which pruning is operated, (2) the criteria uses for selecting parameters to remove, and (3) the schedule followed for pruning. 

\textbf{Granularity. } Pruning granularity is usually divided into two groups: \textit{unstructured} pruning, \textit{i.e.} the weights are removed individually, without any intent to keep structure in the weights \cite{brain_damage,brain_surgeon2}. This method leads to matrices able to reach a high sparsity level, but difficult to speed-up due to the lack of regularity in the pruning patterns. For those reasons, \textit{structured} pruning, \textit{i.e.} removing groups of weights, have been introduced \cite{cpa}. Those structures can include vector of weights, or even kernel or entire filters when pruning convolutional architectures.

\textbf{Criteria. } Early methods proposed to use the second-order approximation of the loss surface to find the least useful parameters \cite{brain_damage,brain_surgeon2}. Later work also explored the use of variational dropout \cite{var_dropout} or even $l_0$ regularization for parameter removal \cite{l0}. However, it has been shown recently that, even though those heuristics may provide good results under particular conditions, they are more complex and less generalizable than computing the $l_1$ norm of the weights, and using that value as a measure of the importance of the weights, thus removing the ones with the lowest norm \cite{state}. Moreover, when comparing the importance of weights, one might do it \textit{locally}, \textit{i.e.} only compare weights that belong to the same layer, which will provide a pruned model that possesses the same sparsity in each layer. One also might compare the weights \textit{globally}, \textit{i.e.} the weights from the whole model are compared when the pruning is applied, resulting in a model with different sparsity levels for each layer. Comparing the weights globally usually provides better results but may be more expensive to compute when the model grows larger in size.

\textbf{Scheduling. } There exist many ways to schedule the network pruning. Early methods proposed to remove weights of a trained network in a single-step, the so-called one-shot pruning \cite{Li}. Such a strategy typically required further fine-tuning of the pruned model in order to recover from the lost performance. However, performing the pruning in several steps, \textit{i.e.} the iterative pruning, is able to provide better results and reach a higher sparsity level \cite{han,iterative}. Nevertheless, such methods were usually time-consuming because of the alternation of many iterations of pruning and fine-tuning  \cite{Li}. Recently, another family of schedules has emerged, performing a pruning that is more intertwined with the training process, allowing to obtain a pruned network in a more reasonable time \cite{to_prune,one_cycle}.

\section{\uppercase{Experiments}}
\label{sec:exp}
 The experimental setup consists in applying unstructured global pruning to the model proposed by \cite{mann} with a hidden layer size $h_{size} = 512$ and 8 experts. As explained in section \ref{subsec:dda}, it is defined by two fully connected-based subnetworks : the gating network extracting the coefficients $\omega_i$ and the motion prediction network whose its weights are the result of the \textit{MoE} processus. The chosen pruning scheduling is the \textit{one cycle} pruning \cite{one_cycle} with the weights \textit{l1} norm criterion \cite{state}. We use the same training data as \cite{mann} which is composed of motion capture recording of dog locomotion with various gaits. 
 \paragraph{}
 First of all, we increase the network sparsity step-by-step from $10\%$ to $90\%$ of the total amount of parameters to extract the relationship between the network sparsity and the motion quality. To quantitatively assess the performance of each model, we measure the foot skating artifact on the generated motion. Foot skating is the fact that the character foot slides when the related joint on the skeleton is considered on contact with the ground which is practically defined if this joint foot height $h$ is under a height threshold $H$. That effect has a bad impact on the motion naturalness and is induced by the mean regression during the training process as explained by \cite{mann}. We use the equation from \cite{mann} to quantify the foot skating $s$ where $v$ stands for the foot horizontal velocity. We fixed the threshold $H$ at $2.5cm$.
\begin{equation}
\label{eq:sk}
    s = v(2-2^{h/H})
\end{equation}

\paragraph{}
Then, we draw a comparison of the generated motion quality between the same size dense (unpruned) and sparse models (pruned). Next, the qualitative contribution of each expert in the generation of the movement is established through an ablation study in order to visualize possible differences between their roles in the case where the initial model is pruned. Finally, the dynamic behavior of the gating network output vector $\omega$ is extracted and subjectively compared between the dense and sparse network. 

For these experiments, we use the \textit{Fasterai} framework \cite{fasterai} built on top of \textit{Fastai} \cite{fastai} to implement the pruning methods. We train 150 epochs on a GTX-1080 Nvidia GPU with a batch size of $32$, a learning rate $\eta=1e-4$ and a weight decay rate $\lambda=2.5e-3$ with a AdamWR algorithm \cite{adamwr} warm restart with the same parameters as \cite{mann}.

\section{\uppercase{Results}}

\label{sec:res}
This section presents the experimental results of the analysis setup described in section \ref{sec:exp}.
\subsection{Performance/Sparsity Analysis}
\begin{figure}[!h]
    \centering
    \includegraphics[scale=0.55]{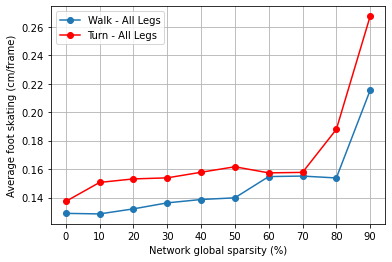}
    \caption{Average Foot Skating on all legs (cm/frame). The generated motion has been evaluated on foot skating while the dog was walking around and abruptly turning.}
    \label{fig:res}
\end{figure}

Figure \ref{fig:res} draws the average foot skating curves along the network sparsity. The generated dog motion is evaluated when he walks and when he performs sudden turns. The trend is that skating increases with the network sparsity. So, there is a trade-off between the number of non-zero parameters and the quality of the synthesized motion. This is due to the fact that the network is less able to learn complex relationships when it has fewer parameters and the generated motion tends towards an average pose which is reflected in the motion by an excess of foot skating. So, following Table \ref{tab:flops} that shows the relationship between the sparsity imposed on the network, the number of non-zero parameters and the floating point operations, the same trade-off exists between the synthesized motion quality and the model size and \textit{FLOPs}. A video showing the degradation of the motion quality when the number of non-zero parameters decreases is available at \textcolor{blue}{\underline{\textit{\href{https://www.youtube.com/watch?v=RHNLQ2Cbz3Y}{https://www.youtube.com/watch?v=RHNLQ2Cbz3Y}}}}.

\begin{table}[!h]
    \centering
    \begin{tabular}{c|c|c}
         Network sparsity & Model size (Mb) & \textit{MFLOPs}\\ 
         \hline
          0\% & 178 & 11.10 \\
          10\% & 160  & 10.06\\
          20\% & 142 & 9.02 \\
          30\% & 124 & 7.98 \\
          40\% & 106 & 6.94 \\
          50\% & 88 & 5.89 \\
          60\% & 71 & 4.86 \\
          70\% & 53  & 3.82 \\
          80\% & 35 & 2.78 \\
          90\% & 17.8 & 1.74\\
    \end{tabular}
    \caption{Relationship between network size and \textit{FLOPs}. Each individual parameter is stored as a 32 bits float.}
    \label{tab:flops}
\end{table}

\subsection{Comparison between same size models}
This section presents a comparison of the quality of the motion generated by two models with an equivalent number of parameters: the size of the dense model hidden layer $h_{size}$ is set to 256, 128 and then 64. Starting from the original network (8 experts and $h_{size}=512$), we prune the network until we reach an equivalent number of parameters for each dense model.
\begin{figure}[!h]
    \centering
    \includegraphics[scale=0.35]{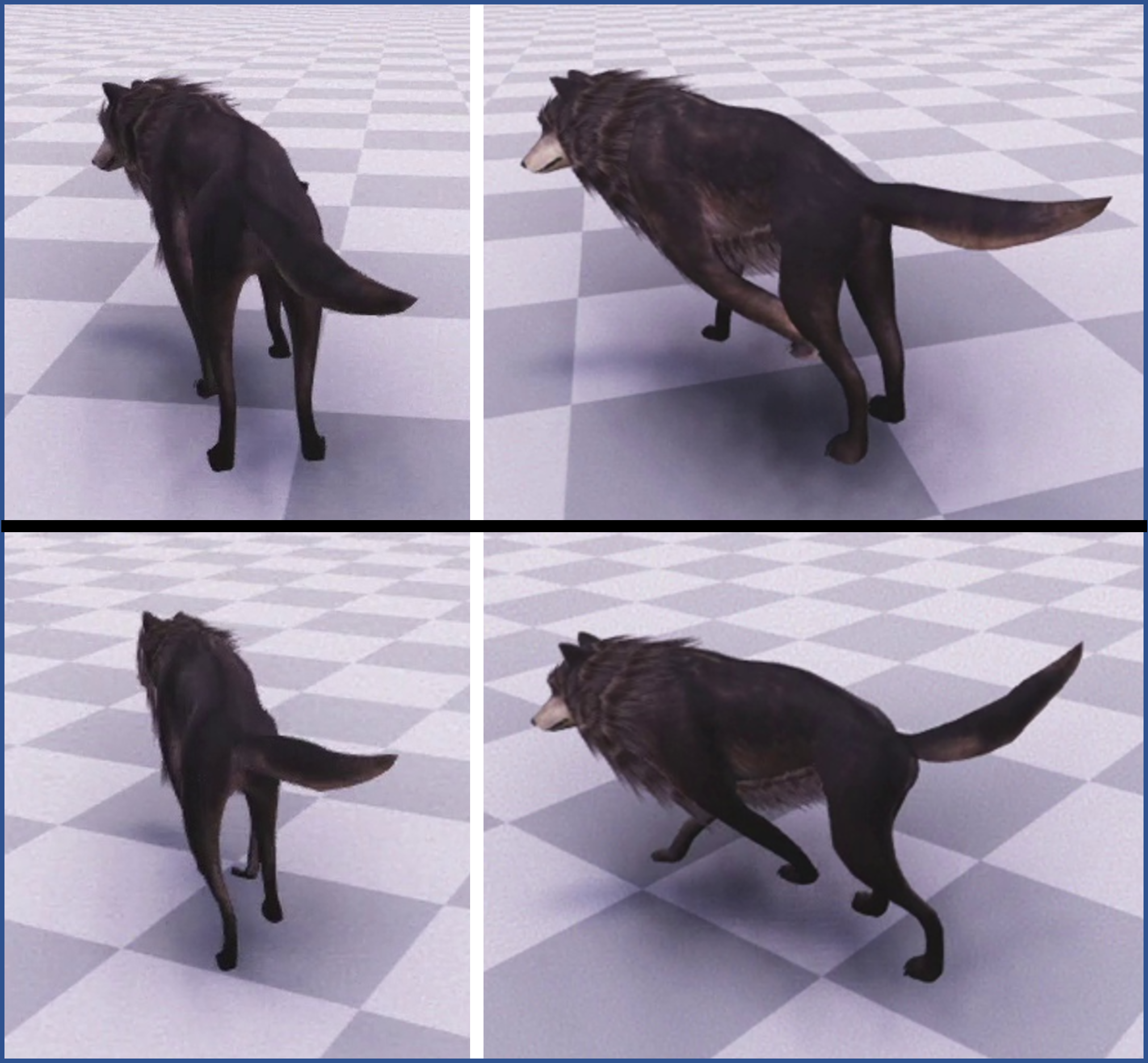}
    \caption{Top : Animation generated by the dense model. - Bottom : Animation generated by the sparse model. Both models have the same number of non-zero parameters. In this example, we keep 10\% of parameters from the initial model. The output motion of the sparse model exhibits less motion artifact than the dense one such as back legs stiffness while turning. The videos about these observations can be found at
    \textcolor{blue}{\underline{\textit{\href{https://youtu.be/AuJ0q019-nc}{https://youtu.be/AuJ0q019-nc}}}}.}
    \label{fig:animation}
\end{figure}

The observations from the related videos show that the quality of motion decreases with the hidden layer size. Indeed, since the network training is a regression problem, it suffers from mean regression and the output poses may converge to a mean pose that minimizes the mean squared error. When reducing the number of parameters, the network has less capacity to learn complex features and it enhances the mean pose regression issue. However, sparse models eliminate globally rough artifacts such as the absence of motion while turning or unnatural poses. This phenomenon is illustrated by Figure \ref{fig:animation} showing an example of the virtual dog animation comparing pruned model with 90\% of sparsity and the equivalent dense model. The dog motion from the dense model is very static while turning.  

\begin{figure}[!h]
    \centering
    \includegraphics[scale=0.55]{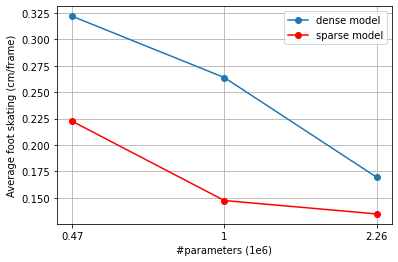}
    \caption{Comparison on average foot Skating on all legs (cm/frame) for dense and sparse models with the same number of parameters and experts. Sparse models exhibit less skating than dense model}
    \label{fig:sk}
\end{figure}

Then, average foot skating is measured for each of the three dense and sparse models while the dog is walking. The abrupt turn motion is not evaluated by skating because the motion artifacts produced by the dense model when reducing $h_{size}$ are too harsh and may not be correctly assessed by Equation \ref{eq:sk}. The results of this evaluation are shown in Figure \ref{fig:sk}. It illustrates the fact that sparse models synthesize motion with less skating than the dense models with the same number of parameters and experts. This points out the benefits of the application of this pruning method on this neural network : while fixing the training budget at 150 epochs in our case, pruning a dense model into a sparse one leads to more reasonable movements than the dense network with the same number of non-zero parameters.

\subsection{Ablation study}
\label{subsec:ablation}

\begin{table*}[!h]
    \centering
    \begin{tabular}{c|c|c}
         $\alpha_i$ & \textit{MANN8} & pruned  \textit{MANN8}\\ 
         \hline
         $\alpha_0$ &  fail to sit and lie &  fail to sit and lie \\
         $\alpha_1$ &  turn badly and fail to lie & turn badly and fail to lie\\
         $\alpha_2$ &  cannot turn while walking & \textbf{walk badly} and cannot turn while walking\\
         $\alpha_3$ &  cannot jump or run &  cannot jump or run and \textbf{turn badly}\\
         $\alpha_4$ &  \textbf{cannot run and badly walk,turn and jump} &  \textbf{can barely move}\\
         $\alpha_5$ &  badly right turn &  badly right turn\\
         $\alpha_6$ &  can only jump &  can only jump\\
         $\alpha_7$ &  cannot jump and run &  cannot jump and run\\
    \end{tabular}
    \caption{Comparison between the ablation study of the dense and the sparse model. Both models exhibit the same experts behavior.}
    \label{tab:m}
\end{table*}

In order to visualize the contribution of each expert to the generation of the movement, we make an ablation study as it is done in \cite{mann} : each expert is deactivated one by one and we observe the impact of this deactivation on the synthesized movement. This study is realized for the initial model with 8 experts and the one pruned with to a sparsity of 90\% (90\% of the total amount of parameters are removed from the network) and is presented in Table \ref{tab:m}. For example, when removing the first expert $\alpha_0$, the dog cannot properly sit or lie on the ground and this behavior is observed on both models. Since each observation while removing the same experts in both models is similar, this ablation study shows that, even in a case of extreme sparsity, each expert in both models corresponds qualitatively to the same high-level features perceived in the movement.  

\subsection{Expert activation behavior}
\begin{figure*}[!htbp]
    \centering
    \includegraphics[scale=0.35]{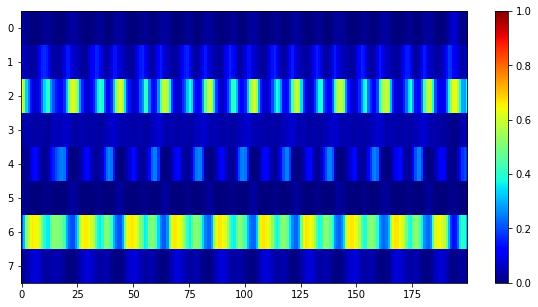}
    \includegraphics[scale=0.35]{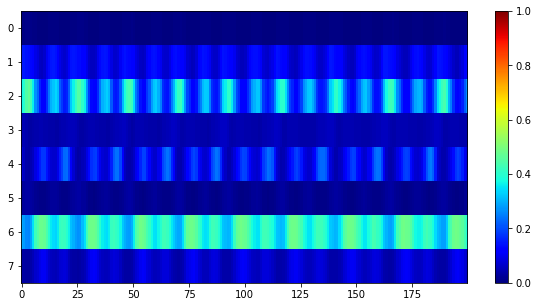}
    \includegraphics[scale=0.35]{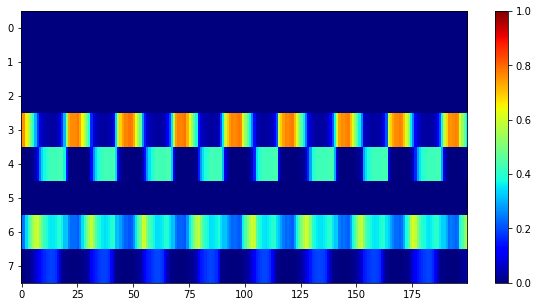}
    \includegraphics[scale=0.35]{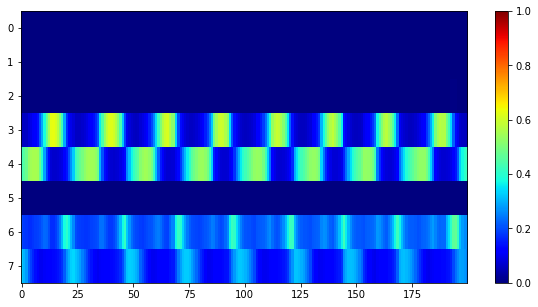}
    \includegraphics[scale=0.35]{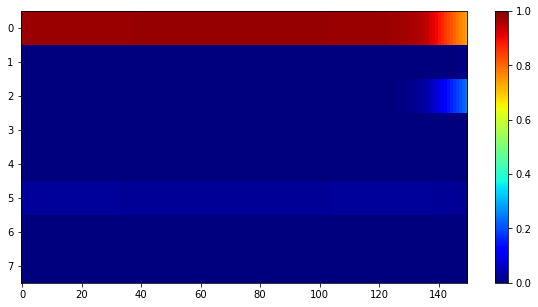}
    \includegraphics[scale=0.35]{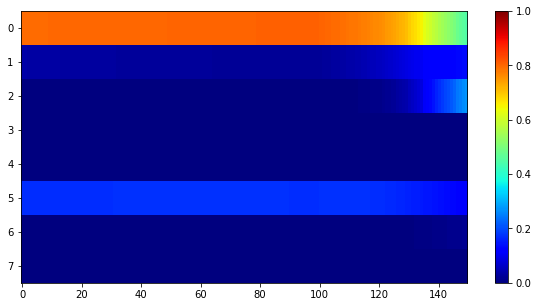}
    \includegraphics[scale=0.35]{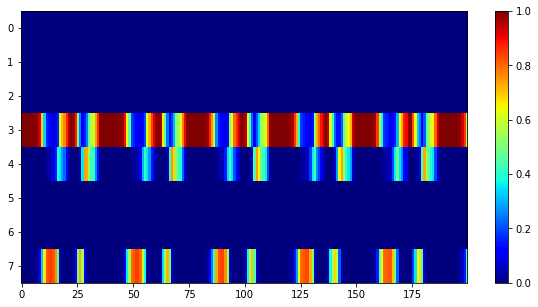}
    \includegraphics[scale=0.35]{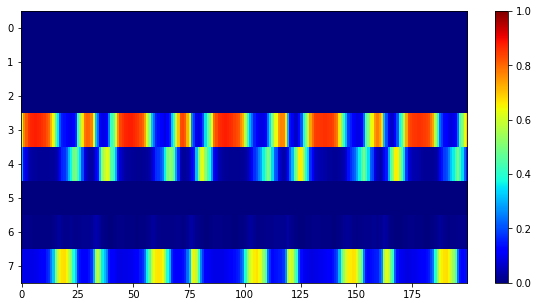}
    \includegraphics[scale=0.35]{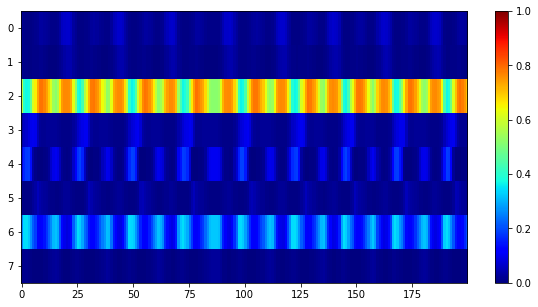}
    \includegraphics[scale=0.35]{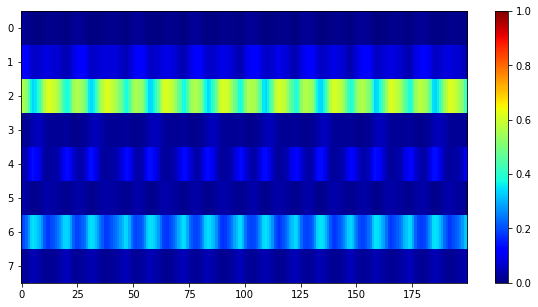}
    \caption{Expert activation $\omega_i$ performing various motion. The left column shows the results for the dense model and the right column for the 90\% sparse model. The action performed from top to bottom: walk, run, sit, jump and turn. A similar behavior is observed for both models: the same experts contribute to the same movements. However, the curves are slightly flatter for the sparse model, which implies that the most significant expert in the dense model is a little less significant in the sparse model.}
    \label{fig:act}
\end{figure*}  

Since section \ref{subsec:ablation} shows that the high-level motion features learned by the experts are not affected by the employed pruning paradigm, the same behavior of the $\omega_i$ (values weighting each expert to compute pose regression network parameters) is thus expected. So, the experts activations are compared between the initial and the sparse model for the same movement. These activations are shown in Figure \ref{fig:act}. When the quadruped performs a specific action, the same weighting profile is applied to experts for both models. However, the activations in the sparse model are slightly more distributed among the different experts compared to the original model. We believe that this is due to the fact that, as the number of non-zero parameters in the pruned model experts is lower than in the dense one, the sparse network exploits those of other experts in order to synthesize similar movements.

\section{\uppercase{Limitations and Perspectives}}

In this work, we applied an unstructured parameter-by-parameter pruning method, \textit{i.e.} with a granularity at the weight level. Using such a method leads, from a dense matrix, to a sparse matrix where no constraint is a priori imposed on the structure of the sparsity of the matrix. This means that, in order to take advantage of the optimization that pruning theoretically brings, it is necessary to have an adequate hardware that can handle sparse matrices. The most evident way to overcome this limitation would be to perform a structured pruning, which would remove blocks of weights, allowing to change the initial architecture, instead of keeping sparse matrices.

Then, The model we considered in our experiments constitutes the backbone of recent models as shown in section \ref{subsec:dda}. One of the perspectives of this work is to be able to prune these models in order to visualize and measure the impact on more complex movements.

Finally, other methods can be employed to optimize a neural network such as knowledge distillation \cite{kd} in which a lightweight student model learns the output of a large master model and tries to reach its performance or fully-connected layers decomposition by singular value decomposition technique \cite{svd} which replaces them by an approximation of two smaller layers. 

\section{\uppercase{Conclusion}}

In this work, we explored the influence of pruning on \textit{MLP-MoE} based architectures, measured and visualized the impact on the generated motion. These analyses showed that:
\begin{itemize}
    \item There is a trade-off between the naturalness of the generated motion and the size of the network and its computational cost.
    \item For an equivalent number of parameters, a pruned network performs better at generating natural motion than a dense network.
    \item Even in the case of extreme sparsity (90\% of the network parameters have been removed), the high level motion features that each expert learns are very similar between the dense and sparse models, which is consistent with the similarity of the activation profiles between these models.
\end{itemize}

As far as we know, this work is the first to make use of pruning techniques in the context of neural network animation and we hope that this work will open the way to further research in the context of lightweight neural animation.

\section*{\uppercase{Aknowledgments}}

Unity materials and dog motion capture files we used here is given by Sebastian Starke \footnote{\href{https://github.com/sebastianstarke/AI4Animation}{https://github.com/sebastianstarke/AI4Animation}}. We would particularly like to thank him for his precious help. 
\bibliographystyle{apalike}
{\small
\bibliography{example}}

\end{document}